\documentclass[10pt,twocolumn,letterpaper]{article}

\usepackage{cvpr}    
\usepackage[math]{alp}
\usepackage{float}
\usepackage{graphicx}
\usepackage{amsmath}
\usepackage{multirow}
\usepackage{amssymb}
\usepackage{booktabs}
\usepackage{soul}
\usepackage{xcolor}
\usepackage{tcolorbox}
\usepackage{algorithm} 
\usepackage{algorithmic} 

\definecolor{cvprblue}{rgb}{0.21,0.49,0.74}
\usepackage[pagebackref,breaklinks,colorlinks,citecolor=cvprblue]{hyperref}

\def\paperID{*****} 
\def\confName{CVPR}
\def\confYear{2024}

\title{Reason out Your Layout: Evoking the Layout Master from Large Language Models for Text-to-Image Synthesis}

\newcommand\blfootnote[1]{%
  \begingroup
  \renewcommand\thefootnote{}\footnote{#1}%
  \addtocounter{footnote}{-1}%
  \endgroup
}
\begin{document}

\author{
    Xiaohui Chen$^\text{1*}$
    \quad
    Yongfei Liu$^\text{2}$ \quad
    Yingxiang Yang$^\text{2}$ \quad
    Jianbo Yuan$^\text{2}$ \quad\\
    Quanzeng You$^\text{2}$ \quad
    Li-Ping Liu$^\text{1}$ \quad
    Hongxia Yang$^\text{2}$ \\\\
    $^\text{1}$Tufts University,~$^\text{2}$ByteDance Inc.\\
    {\tt\small \{xiaohui.chen, liping.liu\}@tufts.edu} \\
    {\tt\small \{liuyongfei.0314, yingxiang.yang, jianbo.yuan, quanzeng.you, hx.yang\}@bytedance.com}
}
\twocolumn[{%
	\maketitle
	\renewcommand\twocolumn[1][]{#1}%
	\begin{center}
		\centering
            \includegraphics[width=\textwidth]{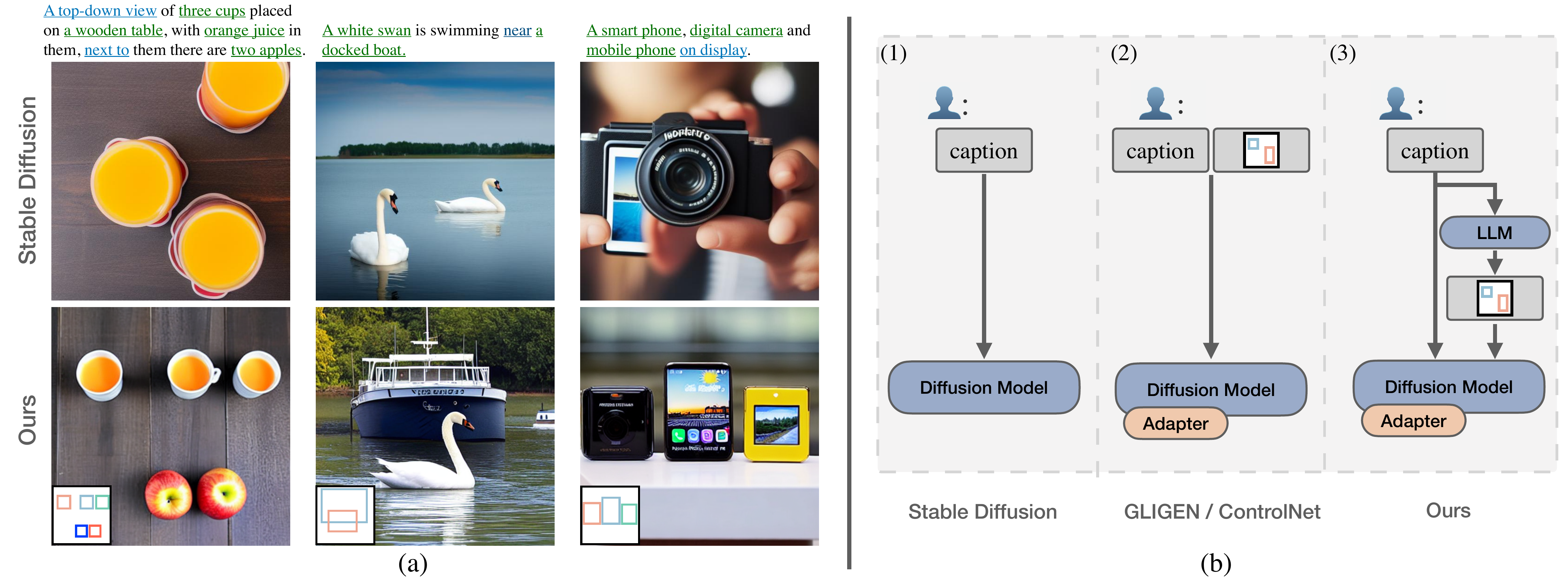}
		\captionof{figure}{(a) Our method enhances the compositional capability of a pre-trained text-to-image diffusion model~\citep{rombach2022high} by conditioning on object layouts; (b) Unlike GLIGEN or ControlNet, which requires manually annotating the layout modality, our method uses LLMs to generate one from the given text prompt.}
		\label{fig:teaser}
		\vspace{0.1in}
	\end{center}
}]
\begin{abstract}

Recent advancements in text-to-image (T2I) generative models have shown remarkable capabilities in producing diverse and imaginative visuals based on text prompts. Despite the advancement, these diffusion models sometimes struggle to translate the semantic content from the text into images entirely. While conditioning on the layout has shown to be effective in improving the compositional ability of T2I diffusion models, they typically require manual layout input. In this work, we introduce a novel approach to improving T2I diffusion models using Large Language Models (LLMs) as layout generators. Our method leverages the Chain-of-Thought (CoT) prompting of LLMs to interpret text and generate spatially reasonable object layouts. The generated layout is then used to enhance the generated images' composition and spatial accuracy. Moreover, we propose an efficient adapter based on a cross-attention mechanism, which explicitly integrates the layout information into the stable diffusion models. Our experiments demonstrate significant improvements in image quality and layout accuracy, showcasing the potential of LLMs in augmenting generative image models.
\end{abstract}

\blfootnote{$^*$ work done during internship at ByteDance.}

\vspace{0em}
\section{Introduction}
\vspace{0em}
\label{sec:intro}

Recent developments in image generation, particularly with DELL-E 2~\citep{ramesh2022hierarchical} and Stable Diffusion (SD)~\citep{rombach2022high}. Specifically, text-to-image (T2I) models, which create images from textual descriptions using autoregressive and diffusion methods, have shown a notable ability to produce high-quality images~\citep{saharia2022photorealistic,ramesh2021zero,ramesh2022hierarchical}. SD-based models, in particular, have obtained significant attention in the research community due to their public availability. However, creating realistic images from complex descriptions still remains challenging. For instance, when dealing with descriptions that include multiple objects with complex spatial relationships, SD-based models often struggle to compose these elements within an image accurately. Fig.~\ref{fig:teaser}(a) shows some examples when multiple objects are described in the text prompt, SD-based models fail to capture them all in the images.

The compositional challenges in SD-based models, including attribute leakage, incorrect attribute binding, omission of objects, or misinterpretation of relationships between objects, are well documented~\citep{wu2023harnessing,feng2022training}. To improve compositional capabilities, a common strategy is to manually provide object positions as model inputs, circumventing the need for the model to infer the layout independently~\citep{zhang2023adding, li2023gligen}. \citet{li2023gligen}, specifically, suggest using bounding boxes to guide image generation, encoding object positions and descriptions into vectors that influence the latent image development via an attention module. Additionally, other research, such as that by \citet{chefer2023attend} and \citet{liu2022compositional}, proposes modifications to the generation process by adjusting the attention mechanism. Additionally, \citet{chefer2023attend, liu2022compositional} propose attention mechanism adjustments, e.g., modifying attention scores to ensure visual features adequately represent each object, thereby reducing object omissions in images. While these layout-augmented methods have been effective, they suffer from the reliance on human-annotated object locations. Furthermore, the integration methods for layout information in these models can be seen as unnatural, as they fail to utilize the spatial details explicitly~\citep{li2023gligen} or heavily alter the image formation process~\citep{liu2022compositional,chefer2023attend}. 

To minimize human intervention in the training process of text-to-image (T2I) generative models, we leverage the capabilities of Large Language Models (LLMs)~\citep{touvron2023llama1,brown2020language} for generating coherent layouts using Chain-of-Thought (CoT) prompting ~\citep{wei2022chain} (Fig.~\ref{fig:teaser}(b)). In particular, we activate LLMs' potential in generating coherent layouts using CoT prompting~\citep{wei2022chain}. The layouts generated by LLMs provide bounding boxes for each object mentioned in the text prompts. For incorporating the LLM-generated layout into the input of SD-based models, we propose an effective adapter that integrates layout information through a cross-attention mask. This adapter explicitly utilizes the spatial details provided by the layout and is designed to align seamlessly with the conditioning mechanisms of SD.


We empirically our method's efficacy in terms of generation quality, layout accuracy, and composition accuracy. Layout generation accuracy accesses the adapter's precision in placing objects within specified bounding boxes. Composition accuracy, a universal metric for T2I models, measures the successful depiction of text-mentioned objects in the generated images. We also explore how different prompting strategies influence layout creation and, consequently, image quality.  We summarize our contributions as follows:
\begin{itemize}
    \item we propose a new pipeline for layout-aware text-to-image diffusion models;
    \item we use LLMs as layout generators and improve their performance via CoT prompting, which elicits reasoning steps in LLMs for more accurate layout generation;
    \item we propose LACA, an adapter designed to incorporate spatial information from the given layouts \textit{explicitly} into the Stable Diffusion models;
    \item our empirical study demonstrates that our proposed LLM-based layout generator can generate layouts that resemble the real ones, thereby enhancing the composition accuracy of the generated images.
\end{itemize}



\vspace{0em}
\section{Related Work}
\vspace{0em}
\label{sec:related-work}

\subsection{Text-to-Image Generation}
\vspace{0em} 

Text-conditional image generation is a key focus in multi-modal learning, with substantial progress in creating realistic images~\citep{goodfellow2014generative,reed2016generative, ding2021cogview,yu2022scaling, rombach2022high,nichol2021glide}. Diffusion models, particularly those introduced by~\citep{sohl2015deep,ho2020denoising}, have gained prominence in this field, thanks to their iterative refinement process and training stability. For instance, \citet{rombach2022high} introduced a latent diffusion model that achieves high performance on minimal computing power, while \citep{nichol2021glide} developed a method for effectively guiding text to create and edit photorealistic images. Despite these advancements, generating photorealistic images from complex text prompts remains a challenge, as highlighted by~\citep{huang2023t2i,cong2023attribute,li2022stylet2i}. Real-world descriptions often involve detailed scenes with complex object interactions, a task that has been approached previously through scene graph parsing~\citep{johnson2018image,zellers2018neural}. However, few studies have managed to generate images that closely match intricate text prompts. Some researchers have suggested conditioning spatial features, like segmentation or bounding boxes, to enhance spatial relation modeling~\citep{li2023gligen, zhang2023adding}. These methods, however, rely on manually created spatial features. For example, \citet{li2023gligen} require manual layout annotation with bounding boxes, and \citet{zhang2023adding} extract the layout (edge map, segmentation, etc) from a template image, and ask the diffusion models to generate a new one conditioning on it. These approaches require special and often manual treatments for layout creation.

\subsection{Chain-of-Thought (CoT) Prompting for LLMs}
\vspace{0em}
In this study, we explore the use of a Large Language Model (LLM) to automatically generate layouts directly from textual descriptions. LLMs, characterized by their transformer-based architecture and enormous size—often comprising hundreds of billions of parameters—include notable examples like GPT-3~\citep{brown2020language}, PaLM~\citep{chowdhery2022palm}, and LLaMA~\citep{touvron2023llama1, touvron2023llama2}. Trained on extensive textual datasets~\citep{shanahan2022talking}, these models exhibit exceptional ability in understanding natural language and performing complex text generation tasks.

The specific technique we use to extract the layout from an LLM is CoT prompting~\citep{wei2022chain}. This is a method that directs LLMs to deconstruct problems into logical steps, facilitating complex reasoning without iterative modifications to the model's parameters. This approach is particularly effective in scenarios with limited example-based learning. It functions by using straightforward instructional sentences to prompt LLMs to process information sequentially~\citep{kojima2022large, wang2023plan}, or by presenting them with a series of examples that illustrate the reasoning process step-by-step~\citep{wei2022chain,zhang2022automatic}. Typically, CoT is implemented in single interactions, where the model generates a continuous chain of reasoning before arriving at an answer. In this work, we leverage the common knowledge the LLMs accumulated during their training processes to generate plausible object layouts of an image. We further use CoT prompting to enhance the accuracy of the LLMs' responses.



\section{Preliminaries}
\subsection{Latent Diffusion Models} 
Our method builds upon the Stable Diffusion model~\citep{rombach2022high}. The Stable Diffusion model consists of an autoencoder and a latent denoising model. In this setup, an encoder, $\calE$, first transforms an image $\bx\in\mathbb{R}^{H\times W\times 3}$ into a latent code $\bz = \calE(\bx)$. 
This latent code is then decoded back to the original image by the decoder, $\calD$, as $\tilde{\bx}=\calD(\calE(\bx))$. The latent code $\bz\in\mathbb{R}^{h\times w\times c}$ can be viewed as a downsampled visual feature map, where $h<H$ and $w<W$. The training process involves first training the autoencoder, followed by the latent denoising model.

Once the autoencoder is trained, a denoising model $\epsilon_\theta$ is trained to generate the latent code $\bz=\calE(\bx)$ starting from a Gaussian noise $\bepsilon\sim\calN(\bzero,\bone)$. The generation process is the reverse of the diffusion process, which progressively adds noise over a series of timesteps $T$~\citep{ho2020denoising}. Specifically, given an image $\bx$ and the corresponding text $y$, the denoising model $\epsilon_\theta$ is optimized using the following training loss:
\begin{align}
    \calL=\mathbb{E}_{\bz\sim\calE(\bx), \epsilon\sim\calN(\bzero,\bone), t}\big[\|\epsilon-\epsilon_\theta(\bz_t, t, c(y)\|^2_2\big].
\end{align}
Here $\bz_t$ is a noisy version of $\bz$ at timestep $t$, and the text information is extracted by a CLIP text encoder $c(\cdot)$~\citep{radford2021learning}. During training, the conditioning text $y$ is occasionally replaced with empty input $\emptyset$ with some probability $p$ to enable classifier-free guidance (CFG)~\citep{ho2022classifier}. The denoising model $\epsilon_\theta$ incorporates the UNet architecture~\citep{ronneberger2015u} with self-attention and cross-attention layers. The cross-attention layers are primarily used to integrate text information.

\begin{figure}[t]
    \centering
    \includegraphics[width=0.9\linewidth]{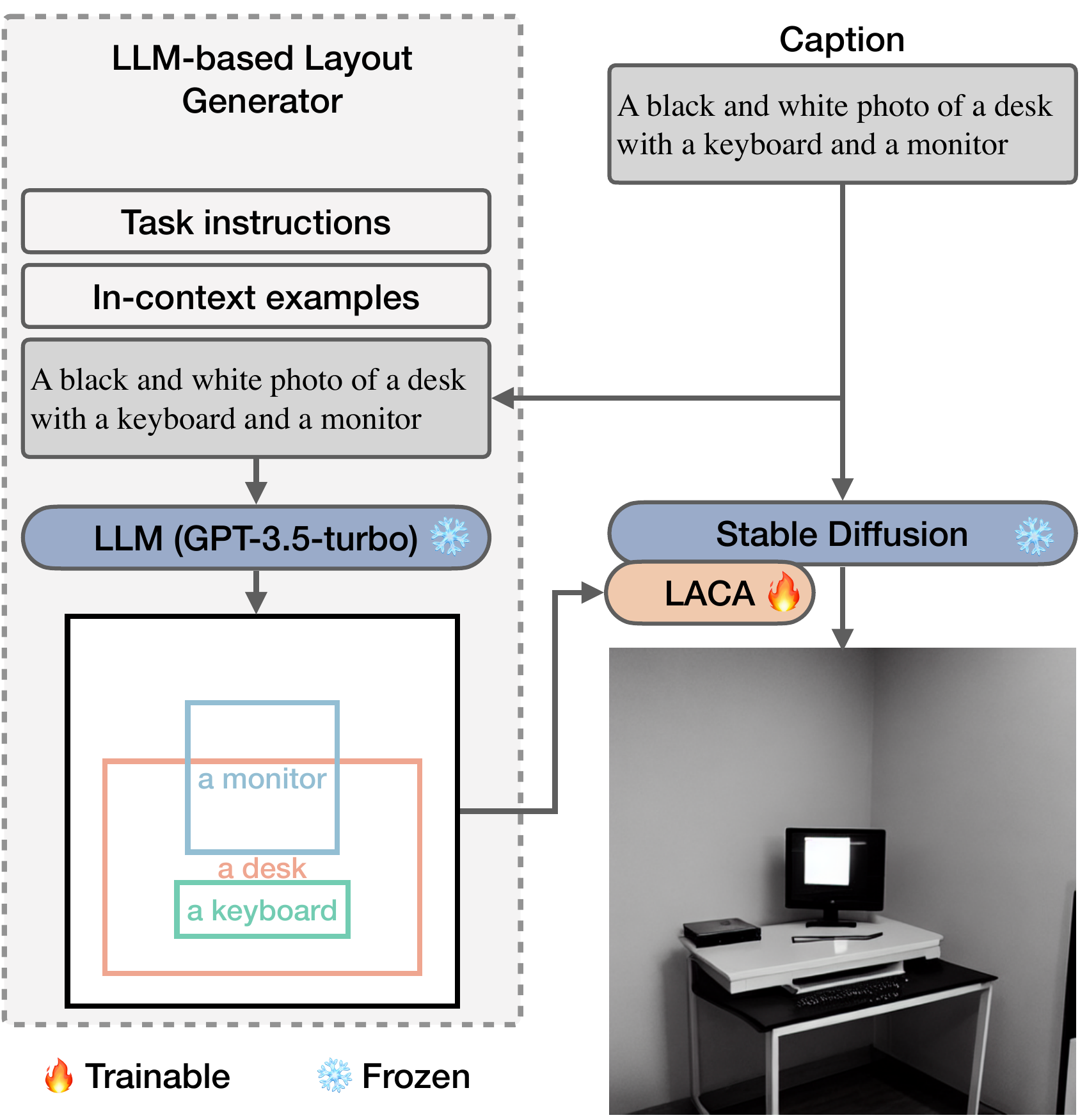}
    \caption{Generation pipeline of our proposed method: Given a caption, we first employ LLM to generate an object layout. This layout is injected into the Stable Diffusion model's noise prediction via our proposed LACA adapter. }\vspace{0em}
    \label{fig:gen-flow}
\end{figure}
\subsection{Injecting Text Modality via Cross-Attention} 
In Stable Diffusion, text information is introduced through cross-attention between the intermediate visual features from the score network $\epsilon_\theta$ and the text embeddings $c(y)$. Let $L$ be the number of downsampling blocks or upsampling blocks in $\epsilon_\theta$, and $p_l:l\in\{1,\ldots, L\}$ be the resolution of the visual feature map output by the $l$-th downsampling blocks.\footnote{For notation simplicity, we focus on downsampling blocks here.} Note that $p_0=h=w$ represents the resolution of the latent code $\bz_t$. Given the text input $y$, its corresponding CLIP text embeddings $c(y)$ are in dimension $\mathbb{R}^{N\times d}$. Here $N$ is the number of tokens and $d$ is the dimension of each token vector. 

For each intermediate visual feature map $\bz_t^{l}$ and the text input $c(y)$, An attention map $\bA_t^l\in\mathbb{R}^{p_l\times p_l\times N}$ is computed using the query (Q) linearly projected from $\bz_t^l$ and the key (K) linearly projected from the text embeddings $c(y)$. Intuitively, in the attention map, each $N$-dimensional slice $\bA_t^l[i,j,:]$ is a probability vector that represents the portion of information each semantic (token) should be aggregated into the visual feature vector at location $(i,j)$ of $\bz_t^l$. The vectors that carry the token semantics are another linear projection (V) of the text embeddings.

\begin{figure*}[t]
    \centering
    \includegraphics[width=1\linewidth]{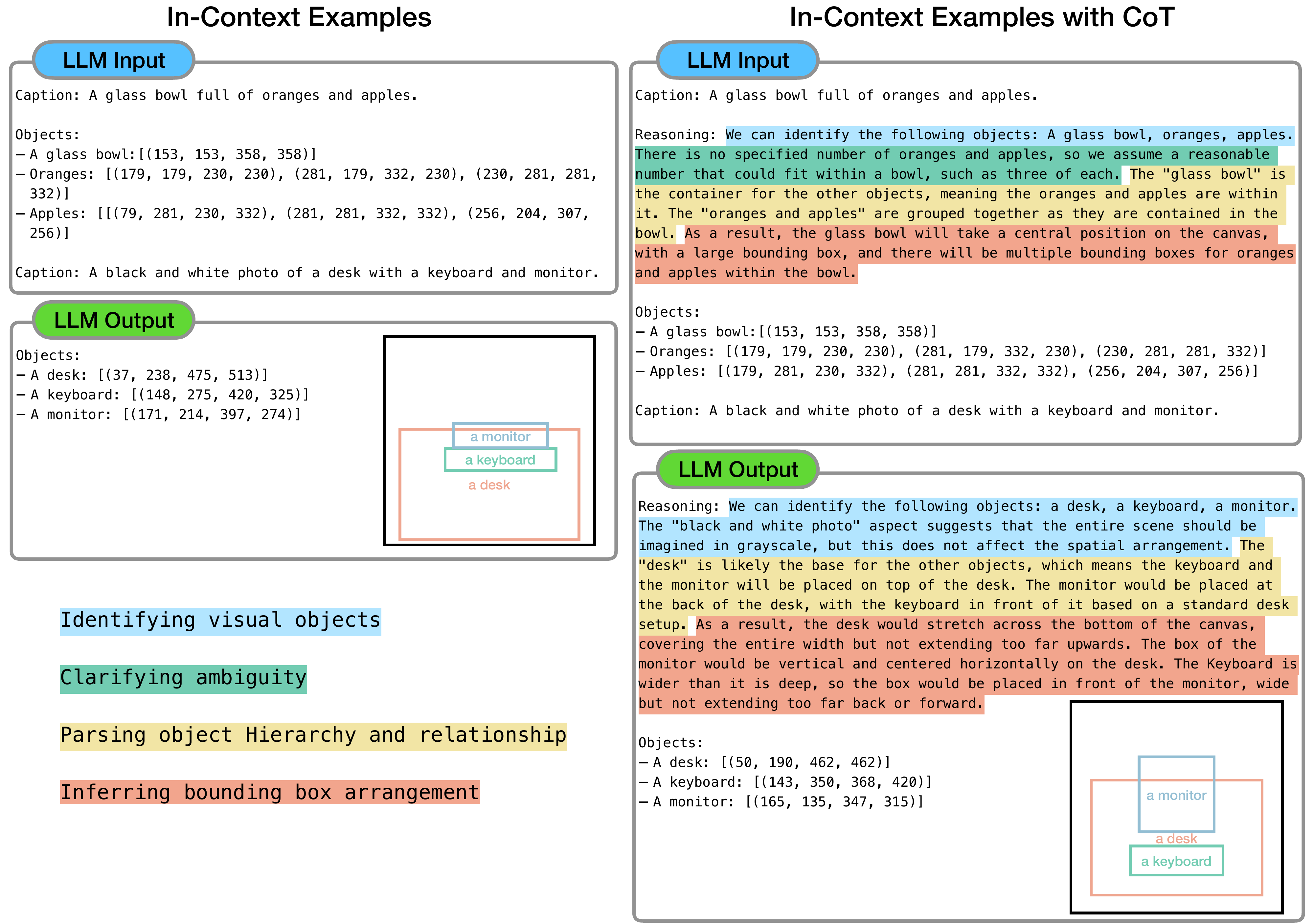}
    \caption{In-context examples with CoT reasoning enable LLM to give a more nuanced layout solution. Reasoning processes of different perspectives are highlighted with different colors. }
    \vspace{-0.5em}
    \label{fig:icl-cot}
\end{figure*}
\vspace{0mm}
\section{Methodology}
\subsection{LLM as Layout Generator}
In this section, we explore the potential of LLMs in generating bounding boxes for objects based on text prompts. Two key factors enable LLMs to generate object layouts: defining \textbf{precise task instructions} and providing \textbf{sufficient in-context examples}. We elaborate on how to turn a general LLM into a layout generator in \S~\ref{sec:task-spec}. In \S~\ref{sec:improved-prompting}, we show how to improve its spatial sense by providing in-context examples with CoT prompting. Fig.~\ref{fig:gen-flow} demonstrates our proposed generation pipeline.

\vspace{-0.5em}
\subsubsection{Task Instructions}
\label{sec:task-spec}
We define clear and comprehensive task instructions for the LLMs to create accurate visual layouts. These instructions focus on distinguishing between visible and abstract elements, resolving ambiguities, understanding spatial relationships, and providing specific object coordinates within a defined canvas area.

\vspace{0mm}
\paragraph{Correctly identifying visual objects.} When parsing objects from the caption, not all objects can be depicted in visual format because they pertain to the other senses or are abstract concepts. For example, in the phrase ``in an office", the entity ``office" represents a scenario composed of multiple objects. Thus, it should be considered as the background of an image rather than an element in object arrangement.

\vspace{0mm}
\paragraph{Resolving ambiguity.} Ambiguities in language, such as vague descriptions or pronouns lacking clear referents, need clarification. For example, pronouns like ``it" or ``they" require specific antecedents to avoid confusion in visual representation. Additionally, exact object quantities should be clarified when descriptions like ``a group of" or ``several" are used.

\vspace{0mm}
\paragraph{Interpreting spatial relations.} Grasping how objects relate to each other in space is crucial. Specifically, recognizing spatial cues such as ``behind", ``in front of", and ``next to" can help generate accurate object placement. Moreover, it is also essential to correctly infer the spatial relationship in the absence of spatial cues. For example, the sentence ``a man holding a tennis racket" implies that the object ``tennis racket" should be placed near the ``man's" hand.

\vspace{0em}
\paragraph{Generating valid answers.} We ask the LLMs to arrange the objects within a $512\times512$ canvas. The top-left coordinate of the canvas is the origin $(0,0)$. For each object, LLMs are asked to provide the bounding box's top-left and bottom-right coordinates. LLMs use the exact phrase from the caption to represent the identified object.

\subsubsection{In-Context Examples with CoT Prompting}
\label{sec:improved-prompting}

In addition to the specific task instructions, we also offer in-context examples to enhance LLMs' ability to produce higher-quality responses. While constructing input-output pairs is straightforward, in this section, we illustrate how to develop CoT prompting to assist the LLMs in generating better outcomes. 

For intricate captions involving several objects, it's advantageous for the LLMs to break down the task into multiple steps and address each sequentially before generating the object layouts. Introducing examples for reasoning in these in-context examples enables the LLMs to emulate such reasoning patterns and engage in CoT reasoning for the caption query. More precisely, we design reasoning frameworks that adhere to the sequence outlined in the task instructions in Section~\ref{sec:task-spec}: (1) identifying both visual and non-visual elements in the caption; (2) resolving any ambiguities regarding the quantity of objects or their attributes, often by making gentle assumptions for clarity; (3) deducing the spatial relations among visual objects in the scene, based on spatial indicators or common knowledge; (4) concluding the arrangement of objects and ultimately deriving the answer from the preceding analysis. Figure~\ref{fig:icl-cot} illustrates how an LLM employs CoT prompting to successfully determine a logical layout, while using standard in-context examples fails to do so.

\begin{table*}[t]
    \centering
    \small
    \begin{tabular}{c l| c c c c c | c c c c c}
        \hline
        && \multicolumn{5}{c|}{\textbf{Flickr30K}}& \multicolumn{5}{c}{\textbf{COCO2017}}  \\
        Layout & 
        \multirow{2}{*}{Method} & 
        \multirow{2}{*}{FID$\downarrow$} &  
        \multicolumn{3}{c}{\colorbox{pink}{YOLO} / \colorbox{yellow}{GLIP} score$\uparrow$} & 
        \multirow{2}{*}{GLIP rate$\uparrow$} & 
        \multirow{2}{*}{FID$\downarrow$} &  
        \multicolumn{3}{c}{\colorbox{pink}{YOLO} / \colorbox{yellow}{GLIP} score$\uparrow$} &
        \multirow{2}{*}{GLIP rate$\uparrow$} \\
        source & & & AP & $\text{AP}_\text{50}$ & $\text{AP}_\text{75}$ &&& AP & $\text{AP}_\text{50}$ & $\text{AP}_\text{75}$\\
        \hline
        - & Stable Diffusion & 22.53 & - & - & - & 78.3 & 20.91 & - & - & - & 71.2 \\
        \hline
        \multirow{2}{*}{GT} & GLIGEN & 27.70 & \colorbox{yellow}{45.0} & \colorbox{yellow}{50.9} & \colorbox{yellow}{45.2} & 84.5 & 27.09 & \colorbox{pink}{19.1} & \colorbox{pink}{30.5} & \colorbox{pink}{20.8} & 71.0 \\
        & LACA & 22.65 & \colorbox{yellow}{50.4} & \colorbox{yellow}{56.6} & \colorbox{yellow}{50.9} & 84.3 & 25.33 & \colorbox{pink}{17.4} & \colorbox{pink}{28.9} & \colorbox{pink}{20.1} & 70.1 \\
        \hline
        \multirow{2}{*}{LLM} & GLIGEN & 30.08 & \colorbox{yellow}{56.2} & \colorbox{yellow}{63.8} & \colorbox{yellow}{56.5} & 83.3 & 25.97 & \colorbox{yellow}{57.6} & \colorbox{yellow}{68.1} & \colorbox{yellow}{46.2} & 70.5 \\
        & LACA & 28.96 & \colorbox{yellow}{58.5} & \colorbox{yellow}{68.1} & \colorbox{yellow}{59.4} & 83.3 & 23.28 & \colorbox{yellow}{58.8} & \colorbox{yellow}{67.3} & \colorbox{yellow}{50.1} & 78.1 \\
        \hline
    \end{tabular}
    \caption{Generative performance on Flickr30K and COCO2017 datasets. FID evaluates image quality, YOLO/GLIP score evaluates correspondence to the conditioned layout, and GLIP rate evaluates composition accuracy. The conditioned layout comes from either the dataset (GT) or the LLM generation.}
    \label{tab:quan-coco}
\end{table*}

\subsection{Stable Diffusion with Layout Conditions}
We propose an attention-based adapter that incorporates the object layout into the score network $\epsilon_\theta$ by manipulating the attention mask $\bM^l\in\{0,1\}^{p_l\times p_l\times N}$. Following~\citet{li2023gligen}, we freeze the original model weight of $\epsilon_\theta$, and introduce a learnable cross-attention module in each transformer block. The original transformer block in $\epsilon_\theta$ consists of a self-attention layer and a cross-attention layer. After adding the layout-aware cross-attention module (LACA), the computation of $\bz_t$ can be written as 
\begin{align}
    &\bz_t = \bz_t+\mathrm{SA}(\bz_t;  \theta);\\
    &\bz_t = \bz_t+\mathrm{LACA}(\bz_t, c(y), \bM; \phi);\\
    &\bz_t = \bz_t+\mathrm{CA}(\bz_t, c(y);\theta).
\end{align}
We omit the superscript $l$ in the following for notation simplicity. We denote $\phi$ to be the parameters of the adapter. In the following, we elaborate on the design details of the proposed adapter. 

\begin{figure}[t]
    \centering
    \includegraphics[width=1\linewidth]{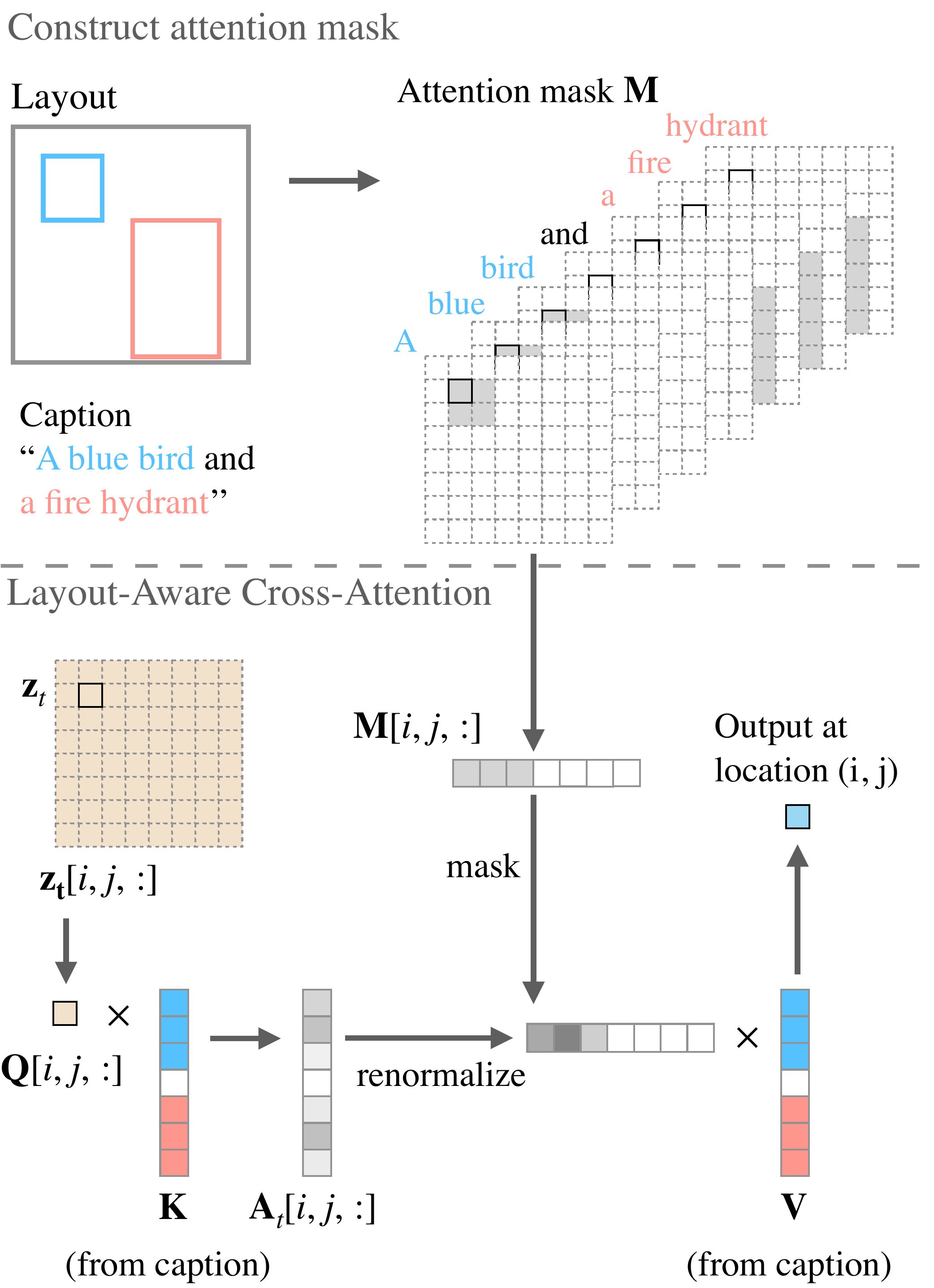}
    \caption{LACA injects layout information via cross-attention mask. The cross attention mask indicates what semantics (e.g., "a blue bird") should attend to the (i,j)-th location (i,j) in the visual feature map. After computing the attention score, the cross-attention mask will enforce the (i,j)-th visual vector only aggregate the semantics from the designated token embeddings.   }
    \label{fig:laca}
\end{figure}
\subsubsection{Layout-Aware Cross-Attention}
Recall that the visual feature vector $\bz_t[i,j,:]$ is updated with the weighted sum of the linear projection (V) of $c(y)$, where the weight is determined by the cross-attention map $\bA_t[i,j,:]$. When the layout indicates the presence of an object in a specific location $(i,j)$ of the feature map $\bz_t$, it is intuitive to aggregate only the semantics of the specific object into this area. 

\paragraph{Composing attention mask from layout.}
Since we specify the LLM to use the same phrase in the caption to represent the object, it is easy to locate the token indices corresponding to an object. Given the token position information and the layout spatial information of each object, we can construct the attention mask $\bM$ with the following rules:
\begin{enumerate}
    \item if the $n$-th token describes an object who is assigned to location $(i,j)$, $\bM[i,j,n] = 1$;
    \item if the $n$-th token does not describe any object, $\bM[i,j,n] = 1$;
    \item otherwise, $\bM[i,j,n] = 0$.
\end{enumerate}
When the $n$-th token is assigned to a specific object, we use rule 1 and rule 3 to decide the attention mask value based on the object's location. When the $n$-th token is not assigned to any object, we set the attention mask $\bM[:,:,n]$ to be $\bone$ such that the token attends to all visual features. Such construction ensures the visual features $\bz_t$ aggregate desired semantics from the text embeddings. Fig.~\ref{fig:laca} provides an example of constructing such an attention mask and how LACA computes a single visual feature $\bz_t[i,j,:]$.

\paragraph{Module designs.} We directly initialize LACA with the model weight from the cross-attention module that follows it in the Stable Diffusion. Moreover, we add an additional zero convolution layer~\citep{zhang2023adding} at the output stage of LACA such that the training is more stable. 

\subsubsection{Sampling}
\label{sec:sampling}
During generation, we only use LACA for the first 20\% denoising steps, then the standard denoising scheme for the rest steps. This is because once the latent code $\bz_t$ encodes a certain amount of the semantics of the objects, the original cross-attention mechanism will function properly. In other words, to improve compositional capability, all an LDM needs is a boost in the early stage of the generation. When using LACA, we employ the classifier-free guidance similar to~\citet{brooks2023instructpix2pix}:
\begin{align}
\tilde{\epsilon} &= \epsilon_\theta(\bz_t,t,\emptyset)+g_1\big(\epsilon_\theta(\bz_t,t,c(y))-\epsilon_\theta(\bz_t,t,\emptyset)\big)\nonumber\\
&+g_2\big(\epsilon_{\theta,\phi}(\bz_t,t,c(y),\bM)-\epsilon_\theta(\bz_t,t,c(y))\big).
\end{align}
Empirically we find such a setting works best for our method. We also investigate other possible CFG options in our experiment (see Appendix~\ref{app:result}).
\begin{figure*}[t]
    \centering
    \includegraphics[width=\linewidth]{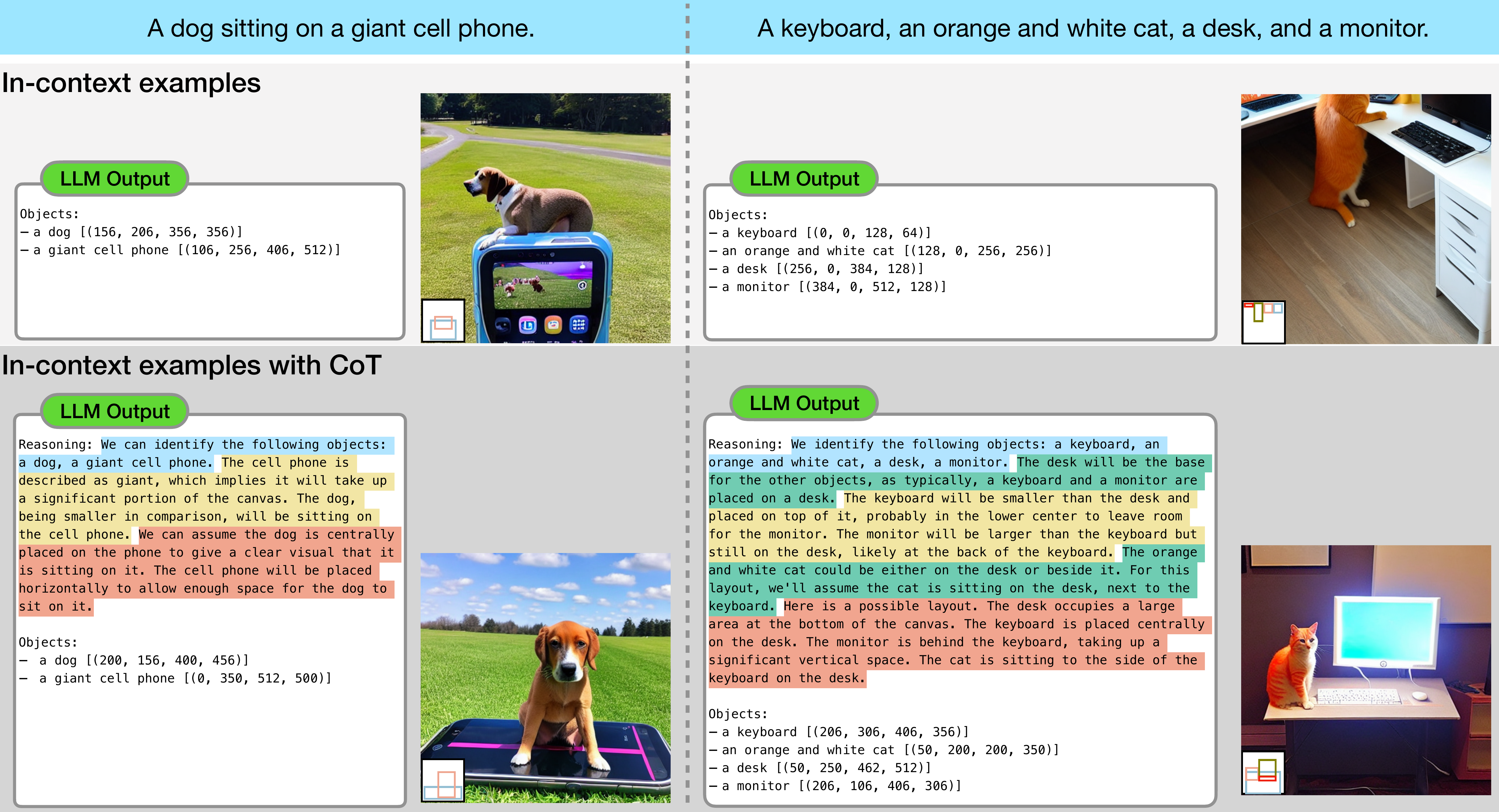}
    \caption{Demonstration of how in-context examples with CoT improve the quality of the generative layout. When the caption describes a counterfactual scene (left) or does not explicitly provide object relation(right), CoT enables LLMs to make assumptions and inferences before generating the layout.}
    \label{fig:enter-label}
\end{figure*}
\section{Experiments}
In this section, we evaluate the quality of the LLM-generated layouts and validate the effectiveness of the proposed adapter. 

\subsection{Setup}
We follow~\citet{li2023gligen} to train the LACA adapter on a combination of four grounding datasets: Object365~\citep{shao2019objects365}, GoldG~\citep{li2022grounded}, CC3M~\citep{sharma2018conceptual} and SBU~\citep{ordonez2011im2text}. We train LACA with batch size 128 and a learning rate 5e-5 for 700k iterations using 8 A100 GPUs. To enable classifier-free guidance, we randomly drop captions with 5\% probability and both caption and layout with another 5\% probability. We set $g_1=g_2=5.5$ during generation for all experiments. We generate all layouts using GPT-3.5-turbo~\citep{chatgpt}.  

\subsection{Quantitative Evaluation on Flickr and COCO}
We first evaluate our method's generative quality and its ability to accurately compose elements using the Flickr30K~\citep{plummer2015flickr30k} and the COCO2017~\citep{lin2014microsoft}. Both datasets are standard benchmarks, in which each image sample comes with caption and object annotations like bounding boxes and semantic masks. We focus on object bounding boxes in our work. A main distinction between the two datasets is that Flickr30K derives the bounding boxes' noun entities directly from the captions, whereas COCO2017's boxes and captions might refer to different objects in the images.

\vspace{0mm}
\paragraph{Baselines.} We use GLIGEN~\citep{li2023gligen} and Stable Diffusion~\citep{rombach2022high} as our baseline methods. We train GLIGEN following the
setting in~\citet{li2023gligen} from the training data. For Stable Diffusion, we use the v1-5 model weight from Huggingface~\citep{wolf2019huggingface}.
\vspace{0mm}
\paragraph{Layout conditions.} We consider two layout conditions - one provided by the datasets, referred to as ground-truth (GT), and one generated by LLMs. The GT layouts in COCO2017 do not match the ground-truth captions and are limited to 80 object classes, whereas the GT layout for Flickr30K and all LLM-generated layouts are matched with their captions. Our proposed method, LACA, only addresses the latter scenario. For COCO2017, with its GT layouts, we modify the captions to include all noun entities present in the layout, making them compatible with LACA.

\paragraph{Evaluations.} We evaluate the models in terms of image quality and layout accuracy. Specifically, we use FID~\citep{heusel2017gans} for image quality, and for layout accuracy, we consider using {YOLO score}~\citep{li2021image} for independent layout condition (COCO2017 with GT) and {GLIP score}~\citep{li2023gligen} for derived layout condition~\citep{li2023gligen}. We report average precision (AP) for both YOLO and GLIP scores. While layout accuracy only applies to layout-conditioned models, we additionally evaluate the composition accuracy of any T2I models. Specifically, the composition accuracy measures how many objects mentioned in the text are present in the generated image. This can be quantified by again using GLIP~\citep{li2022grounded}. Since GLIP derives all entities from a caption and tries to detect the entities in its corresponding images, we can quantify composition accuracy by computing the ratio:
\begin{align}
    \frac{\sum_i \text{\# entities detected in image}~i}{\sum_i\text{\# entities derived from caption}~i},
\end{align}
dupped {GLIP rate}. The numerator is always smaller than the denominator as we only consider known entities. All metrics composition-related metrics ({YOLO score}, {GLIP score}, and {GLIP rate}) are scaled by 100.

\paragraph{Results.} Table~\ref{tab:quan-coco} demonstrates the performance of different models when combined with different layout sources. We interpret the numbers with the following perspectives: (1) By comparing the performance of different models under the same layout source, we can see that LACA consistently achieves better FID and GLIP scores. The difference is more prominent in the Flickr30K dataset. While in the COCO2017 dataset paired with GT layouts, LACA works slightly falls short of GLIGEN regarding the YOLO score. We hypothesize this is because the adapted captions for LACA are not as coherent as the original captions. When compared to Stable Diffusion, both the GLGIEN and LACA demonstrate superior GLIP rates, indicating they are more effective at incorporating objects mentioned in the captions into the generated images. (2) By comparing the performance of the same models under different layout sources, we can observe that in Flickr30K, the FID scores from LLM-generated layouts are worse than from the GT layouts, while in COCO2017, the FID scores from LLM-generated layouts are better. This discrepancy likely stems from the text-layout consistency of the datasets -- a text-consistent layout ease the burden of the diffusion models to integrate objects from multiple modalities.


\begin{table}[t]
    \centering
    \begin{tabular}{l|ccc}\hline
        Prompting strategy &  hit rate & mIoU &FID\\\hline
        Task instruction only &  92.2\% & 16.54 &32.43\\
        + ICL & 97.1\%&  19.84 & 31.95\\
        + ICL w/ CoT   &97.4\%&  23.98 & 28.96\\\hline
    \end{tabular}
    \caption{In-context examples with CoT help improve the layout performance, thus the generated image quality. ICL stands for in-context learning.}
    \label{tab:miou}
\end{table}
\vspace{0mm}
\subsection{Prompting Strategies vs Layout Performance}
\label{sec:exp-cot}
We investigate the effect of different prompting approaches on layout generation using the Flickr30K dataset. Our focus is to compare the resemblance of LLM-generated bounding boxes to the original GT bounding boxes.

\paragraph{Evaluation protocol.} We randomly sample 5000 caption-layout pairs from the dataset. For each caption, an LLM is used to generate corresponding layouts using different prompting strategies. Given the GT layout and the LLM-generated layouts, we measure two metrics: (1) the {object hit rate}, which measures how many objects in the GT layouts have been identified by the LLMs; (2) the {intersection over union (IoU)} of the GT and the LLM-generated bounding boxes. While the generated boxes can be completely different from the ground-truths for an individual comparison, the mIoU is still a reliable measure of the goodness of a layout generator. Additionally, we assess the image quality produced from LLM-generated layouts using the FID score, with a lower FID score indicating better layout quality.

\paragraph{Prompting variants.} We study three types of prompting strategies: (1) task instructions only; (2) task instructions followed by eight in-context examples, but without CoT reasoning; (3) task instructions followed by eight in-context examples, each come with a CoT reasoning.

\paragraph{Results.} Table \ref{tab:miou} reports and mIoU and the object hit rate of the generated layouts. As we can observe, when providing in-context examples, the object hit rate reaches almost 100\% and no longer increases. Moreover, when CoT reasoning is employed, we achieve the highest mIoU, indicating the generated layouts are most similar to the real ones.

\subsection{Generative Counting Accuracy}
\begin{figure}
    \centering
    \includegraphics[width=0.8\linewidth]{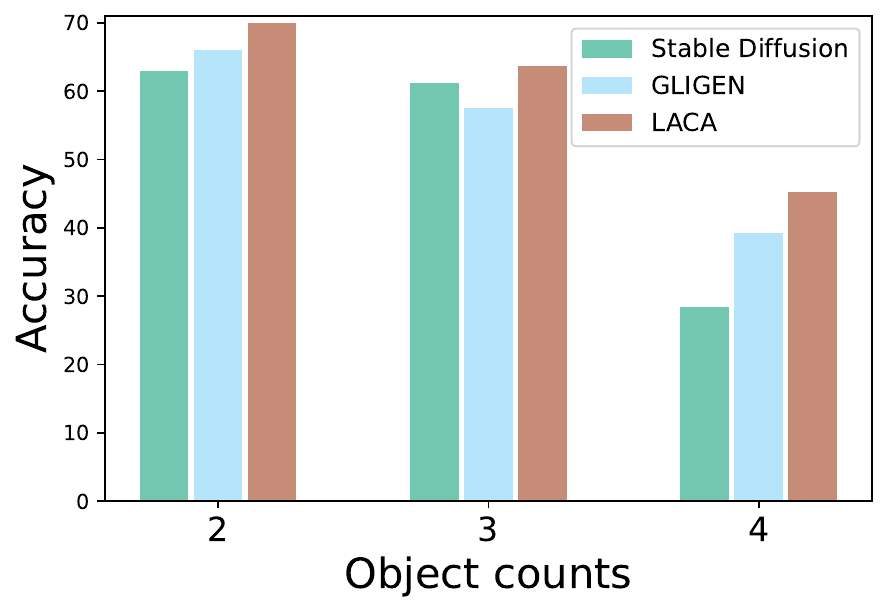}
    \vspace{0mm}
    \caption{LACA achieves better generative counting accuracy compared to GLIGEN and Stable Diffusion. }
    \label{fig:count}
    \vspace{0mm}
\end{figure}

We evaluate the performance of different models in generative counting accuracy, which is their ability to replicate the exact number of objects as described in the text. For example, the model accuracy in generating an image with ``four apples'' when prompted by the text. For this purpose, we curated a set of captions from the COCO2017 validation set that specifically include the numeral keywords ``two'', ``three'' or ``four'', in which 150 captions were collected for each number category. Using LLMs, we generate object layouts for the collected captions. From Fig.~\ref{fig:count}, we can see that LLM-generated layout generally improves the model's accuracy in depicting the precise count of objects. The improvement becomes more notable when the object count rises. Moreover, we can observe that LACA surpasses GLIGEN in its effectiveness in employing the layout modality.

\subsection{Ablation Study on Classifier-Free Guidance}
We explore how the hyperparameters $g_1,g_2$ from classifier-free guidance affect the quality of the generated images. We explore $g_1\in\{1.5,3.5,5.5,7.5\}$ and $g_2\in\{3.5,5.5,7.5\}$ and report the FID scores in Table~\ref{tab:cfg-fid}. We can observe that $g_1=g_2=5.5$ yields the best FID score. 
\begin{table}[h]
    \centering
    \begin{tabular}{lc|cccc}\hline
    &&\multicolumn{4}{c}{$g_1$}\\
        &&  1.5 & 3.5 & 5.5    & 7.5 \\\hline
     \multirow{3}{*}{$g_2$}   &3.5& 27.11 & 26.34 & 26.78 & 28.16  \\
        &5.5 & 26.45 & 23.80 & 23.28 & 23.28\\
        &7.5  & 24.92 & 23.67 & 23.35 & 23.30 \\\hline
    \end{tabular}
    \caption{Classifier-free guidance weights over text and text-layout conditions. $g_1$ controls the text-layout condition and $g_2$ controls the text-only condition.}
    \label{tab:cfg-fid}
\end{table}

\vspace{-1em}
\section{Conclusion and Discussion}
In this work, we propose to use LLMs as layout generators to generate object layouts from the given captions. By introducing the CoT prompting and carefully designing the reasoning steps for the in-context examples, we enable the LLMs' to generate more nuanced layout solutions and showcase the potential of LLMs in understanding and generating complex visual layouts. We further propose LACA, an adapter that explicitly incorporates the object layout information into the Stable Diffusion. We demonstrate that LACA is superior in yielding high composition accuracy, without conditioning on human-annotated layout modality. This work provides clear direction for ongoing research to refine the synergy between textual descriptions and visual generation, paving the way for more sophisticated and accurate visual content creation by LLMs.


\small \bibliographystyle{ieeenat_fullname} \bibliography{main}
\newpage

\begin{table*}
    \centering
    \scriptsize
    \begin{tabular}{lc}
    \hline
        Task instructions &
        \multicolumn{1}{p{14cm}}{You are an expert photographer who can infer the best layout of the given objects inside a photo or a picture. Now, given a description of the picture, you are asked to perform the following tasks.
        
\vspace{0.8em}
1. Given the description, parse the objects that appear in the text in a hierarchical manner.

2. Based on your parsed result, arrange the objects within a canvas with a width of 512 and a height of 512. The top-left coordinate in the canvas is the origin (0, 0).

3. For each object, you need to specify its location by listing the top-left coordinate and the bottom-left coordinate. Your answer for each object should be (x1, y1, x2, y2), where (x1, y1) is the top-left coordinate and (x2, y2) is the bottom-right coordinate.

4. In the description, if there is any ambiguity about the number of objects or the spatial relationship between objects, you should first concretize it through reasoning before giving the answer.

5. When representing the identified objects in your answer, you should use the exact same words that appear in the caption.

Below are a few examples:}\\\\\hline\\
In-Context Example 1 & \multicolumn{1}{p{14.1cm}}{
\#\# \textbf{Caption: A man in a white shirt and blue shorts swinging a tennis racket.}

\vspace{0.8em}
\#\#\# Parsing the description into objects

~~~~From this caption, we can identify the following objects: A man, A white shirt, Blue shorts, A tennis racket.

\vspace{0.8em}
\#\#\# Hierarchy and relationships

~~~~- ``A man'' is the main subject, and his clothing (a white shirt and blue shorts) is part of his description.

~~~~- ``A tennis racket'' is being swung by the man, so it will be in motion, likely extending from one side of the man.

\vspace{0.8em}
\#\#\# Arranging objects on the canvas

~~~~- The canvas is in a size of width and height of 512, with the origin at the top-left (0, 0).

~~~~- The man should be centrally located to be the focus, with space around him to show the movement of swinging the racket.

~~~~- The clothing (white shirt and blue shorts) is part of the man's bounding box, with the shirt on the upper part of the torso and the shorts below.

~~~~- The tennis racket, since in motion, should extend out from the man's hand, likely to the right side if we imagine the swing.

\vspace{0.8em}
\#\#\# Reasoning and concretizing ambiguity

~~~~- The exact positions of the shirt and shorts within the man's bounding box are based on their natural position on the body.

~~~~- The tennis racket's position is determined by the typical posture of swinging, which generally extends to the side and slightly upward.

\vspace{0.8em}
\#\#\# Specifying locations

~~~~- ``A man'' will have a bounding box that covers a significant portion of the canvas to show his presence and the action.

~~~~- ``A white shirt'' will have a bounding box within the upper half of the man's bounding box, representing the torso area.

~~~~- ``Blue shorts'' will have a bounding box below the shirt's, indicating the lower part of the torso and upper legs.

~~~~- ``A tennis racket'' will have a bounding box that overlaps with the man's hand and extends outward to represent the swing.

\vspace{0.8em}
\#\#\# Answer

~~~~- **A man**: visual [[158, 51, 337, 404]]

~~~~- **a white shirt**: visual [[204, 153, 317, 256]]

~~~~- **blue shorts**: visual [[220, 235, 327, 307]]

~~~~- **a tennis racket**: visual [[153, 46, 235, 143]]}\\\\
\multicolumn{1}{p{2.2cm}}{In-Context Example 2

...}& ... \\\\\hline\\
Query & \multicolumn{1}{p{14cm}}{Now given the caption below, can you give a similar reasoning and derive the resulting bounding box for those objects? then give the answer, strictly following the format of the answer given in the examples.

\vspace{0.8em}
\#\# \textbf{Caption: a red apple and a blue bird.}

\vspace{0.8em}
\#\#\#
}
\\\hline
\end{tabular}
\caption{Prompts for LLMs to generate layout for caption ``a red apple and a blue bird.''.}
\vspace{-1em}
\label{tab:example-prompt}
\end{table*}
\appendix

\begin{table*}
    \centering
    \scriptsize
    \begin{tabular}{ccc}
    \hline\\
        \multicolumn{3}{p{14cm}}{\textbf{\#\# Caption: A glass bowl full of oranges and apples. }}\\\\\hline\\ \textbf{CoT variant 1} & \textbf{CoT variant 2} & \textbf{CoT variant 3}\\\\\multicolumn{1}{p{5.1cm}}{
        
\#\#\# Identifying Objects

~~~~From this caption, we can identify the following objects: A glass bowl, oranges, and apples.

\vspace{0.8em}
\#\#\# Specifying Locations

~~~~- The glass bowl, being the central object, will have a bounding box in the middle of the canvas, perhaps taking up a significant area but not touching the edges to allow for visual clarity.

~~~~- The oranges and apples will each have their own bounding box within the bowl. Since they are grouped together, their boxes may overlap or be side by side.

        }& \multicolumn{1}{p{5.1cm}}{

\#\#\# Identifying Objects

~~~~From this caption, we can identify the following objects: A glass bowl, oranges, and apples.

\vspace{0.8em}
\#\#\# Hierarchy and Relationships

~~~~- The glass bowl serves as the container for the oranges and apples.

\vspace{0.8em}
\#\#\# Arranging objects on the canvas

~~~~- Canvas Size: 512x512 square with the origin at the top-left (0, 0).

~~~~- Bowl Placement: Centrally on the canvas to emphasize its role as a container.

~~~~- Fruit Placement: Oranges and apples inside the bowl, possibly overlapping or side by side.

\vspace{0.8em}
\#\#\# Reasoning and concretizing ambiguity

~~~~- Quantity of Fruit: Assuming a reasonable number, such as three oranges and three apples.

~~~~- Fruit Arrangement: Random scattering within the bowl.

\vspace{0.8em}
\#\#\# Specifying Locations

~~~~- The glass bowl, being the central object, will have a bounding box in the middle of the canvas, perhaps taking up a significant area but not touching the edges to allow for visual clarity.

~~~~- The oranges and apples will each have their own bounding box within the bowl. Since they are grouped together, their boxes may overlap or be side by side.
}& \multicolumn{1}{p{5.1cm}}{We can identify the following objects: A glass bowl, oranges, apples.
There is no specified number of oranges and apples, so we assume a reasonable number that could fit within a bowl, such as three of each. The "glass bowl" is the container for the other objects, meaning the oranges and apples are within it. The "oranges and apples" are grouped together as they are contained in the bowl. As a result, the glass bowl will take a central position on the canvas, with a large bounding box, and there will be multiple bounding boxes for oranges and apples within the bowl.
}\\\\\hline\\
\multicolumn{3}{p{14cm}}{
\#\#\# Answer

~~~~- **A glass bowl**: visual [[153, 153, 358, 358]]

~~~~- **Oranges**: visual [[179, 179, 230, 230], [281, 179, 332, 230], [230, 281, 281, 332]]

~~~~- **Apples**: visual [[179, 281, 230, 332], [281, 281, 332, 332], [256, 204, 307, 256]]}
\\\hline
\end{tabular}
\caption{An in-context example with different CoT variants.}
\label{tab:cot-variant-demo}
\end{table*}

\section*{Appendix}
\section{Implementation Details}
\subsection{Prompt Designs}
We first elaborate on the realization of task instructions, and then we explore different variants of chain-of-thought prompting provided in the In-Context examples.
\paragraph{Task instructions.} The task instructions implemented are shown in Table~\ref{tab:example-prompt}. While most of the instructions are straightforward, we specifically focus on the different ways to depict object placement in the responses from the LLMs. We explore variations from two perspectives: (1) using a normalized 1x1 canvas as opposed to a more expansive 512x512 canvas, and (2) two methods of representing object locations - either through top-left and bottom-right coordinates (XYXY) or by indicating the top-left coordinate with width and height descriptions for the bounding box (XYWH). Our ablation study, which assesses the quality of images produced under these different settings via the FID score, revealed a slightly better performance with the combination of a 512x512 canvas and the XYXY method for bounding box representation.


\paragraph{In-Context Examples.} We explore the three formats of CoT reasoning. The first CoT variant includes a two-step reasoning process: first interpreting the visual objects from text, then creating their arrangements. This basic approach generally suffices for accurate layout creation. However, for cases with complex or vague object relationships, it's advantageous for LLMs to engage in inference and assumption before solution generation. In the second variant, we integrate clarification steps within the CoT reasoning of the first version. The clarification steps are mostly useful for prompts that do not carry enough information on the object position or specification information, such as e.g., quantity, attribute, etc. For the first two variants, each reasoning step in the first two formats is separated by bullet points in order to make it easier for the LLMs to mimic. The final CoT format adheres to the reasoning approach of the second variant but combines all reasoning steps continuously without explicitly delineating each reasoning phase. We provide an example demonstrating those variants in Table~\ref{tab:cot-variant-demo}

 
\subsection{Layout Conversion from LLMs Response}
We extract bounding boxes for objects identified in the responses from LLMs.  These bounding boxes are then transformed into the cross-attention mask $\bM^l$ used by LACA. Without loss of generality, we denote $\bM:=\bM^0\in\mathbb{R}^{64\times64\times N}$ and use $\bM$ for the discussions below.
.
\paragraph{Bounding boxes extraction.} As stated in our task instructions, the canvas that an LLM operates on is a size 512x512. An LLM will first parse objects from the caption and then classify them into either ``visual'' or ``non-visual''. Only "visual" objects are assigned bounding boxes, which are determined by their top-left and bottom-right coordinates. If an object is mentioned in multiple quantities, the LLM will generate a corresponding number of bounding boxes, based on either the specified quantity or its own estimations. After extraction, these coordinates are normalized to a scale ranging from 0 to 1.

\paragraph{Cross-attention mask construction.} For each object, we first locate its indices in the text caption in order to properly assign values to $\bM$. For example, in the text prompt ``a red apple and a blue bird'', the indices of the object ``a red apple'' will be $[0,1,2]$ (assuming the index starts from 0). In practice, since Stable Diffusion uses a CLIP model to encode the prompt, we find such correspondence after the text prompt and the object description are both tokenized. Suppose we have ``a red apple'' matched the $(i,i+1,\ldots,j)$ tokens in the tokenized text prompt, and the top-left coordinate being $(x_1,y_1)$, the bottom-right coordinate being $(x_2,y_2)$, we then will set
\begin{align}
    \bM\big[\lfloor 64x_1\rfloor:\lfloor 64x_2\rfloor, \lfloor 64y_1\rfloor:\lfloor 64y_2\rfloor, i:j\big] = 1.
\end{align}
Note that $\bM$ is initialized with all zeros before composing the layouts on it. $\lfloor\cdot\rfloor$ is the floor operation. For any token index $i$ that does not represent an object, we set the cross-attention mask $\bM[:,:,i]=1$.
\subsection{Classfier-Free Guidance}
Recall that in \S~\ref{sec:sampling}, while we estimate the predictive noise using
\begin{align}
\tilde{\epsilon} &= \epsilon_\theta(\bz_t,t,\emptyset)+g_1\big(\epsilon_\theta(\bz_t,t,c(y))-\epsilon_\theta(\bz_t,t,\emptyset)\big)\nonumber\\
&+g_2\big(\epsilon_{\theta,\phi}(\bz_t,t,c(y),\bM)-\epsilon_\theta(\bz_t,t,c(y))\big),\label{eq:cfg1}
\end{align}
we also explore other possible choices. The first alternative is the one used by~\citet{li2023gligen}, which jointly considers the text and the layout modality and has
\begin{align}
\tilde{\epsilon} = \epsilon_{\theta,\phi}(\bz_t,t,\emptyset)+g\big(\epsilon_{\theta,\phi}(\bz_t,t,c(y),\bM)-\epsilon_{\theta,\phi}(\bz_t,t,\emptyset)\big).\label{eq:cfg2}
\end{align}
For this alternative, we drop both modalities at the same time with a probability of 10\% during the training. The second alternative regarding the choice between using score networks $\epsilon_\theta$ and $\epsilon_{\theta,\phi}$. Intuitively, one should use $\epsilon_{\theta,\phi}$ for all combination of input modality, which yields
\begin{align}
\tilde{\epsilon} &= \epsilon_{\theta,\phi}(\bz_t,t,\emptyset)+g_1\big(\epsilon_{\theta,\phi}(\bz_t,t,c(y))-\epsilon_{\theta,\phi}(\bz_t,t,\emptyset)\big)\nonumber\\
&+g_2\big(\epsilon_{\theta,\phi}(\bz_t,t,c(y),\bM)-\epsilon_{\theta,\phi}(\bz_t,t,c(y))\big).\label{eq:cfg3}
\end{align}
However, we empirically found that the setting in Eq.~\ref{eq:cfg1} works slightly better than Eq.~\ref{eq:cfg3} in terms of generated image quality. And both of Eq.~\ref{eq:cfg1} and Eq.~\ref{eq:cfg3} demonstrate better performance than the one using Eq.~\ref{eq:cfg2}. We report the FID score of those settings in the experiment proposed by Appendix~\ref{sec:cfg-ablation}.

\subsection{Details for mIoU Computation}
In \S~\ref{sec:exp-cot}, we measure how the LLM-generated layouts resemble the ground-truths via mIoU. The mIoU score directly computed between two sets of bounding boxes is extremely low since first, the object labels are open-set, and second, there are many possible layouts for a given caption. To better match the bounding boxes, we make two modifications to the original matching algorithms to increase the reliability of the metric. 
\paragraph{Relaxed object matching.} While current tools for mIoU computation find correspondence between objects from the GT and LLM-generated layouts by matching their noun entities, we build such correspondence by checking whether one is a substring of the other. For example, while the GT layouts describe the phrase ``a woman in a blue shirt'' with one bounding box, an LLM might provide two bounding boxes for ``a woman'' and ``a blue shirt'' respectively. Our approach relates both the bounding boxes of ``a woman'' and ``a blue shirt'' to the GT bounding box for ``a woman in a blue shirt''. 
\paragraph{Bounding box flipping.} For each image, we compute two sets of IoU values, one with the GT bounding boxes and one with the horizontally flipped ones. We then take the set that has a higher mIoU score as the result. We only perform horizontal flips on the bounding boxes since it does not change the spatial sense of an image. 

A proper metric used to measure the layout generative performance under the open-set setting is still underexplored. While our proposed method is shown effective, there might exist a principal solution to such a problem. We would like to leave it to the future work.

\subsection{Model Size}
We list the number of model parameters in Table~\ref{tab:params}. The architecture of adapter LACA+LASA is detailed in Appendix~\ref{sec:lasa}. Our proposed adapters have significantly fewer parameters compared to GLIGEN.

\section{Layout-Aware Self-Attention Module}
\label{sec:lasa}
Inspired by LACA, we further investigate a possible variant of the proposed adapter - Layout-Aware Self-Attention Module, dupped LASA. In this section, we first demonstrate the development of LASA, then we show how to jointly compose the LACA and LASA adapters in the Stable Diffusion model.
The proposed LASA adapter aims to make sure an object has coherent visual features during the generation. The object coherence is enforced by the layout modality -- visual features that belong to the same object should self-attend to each other. Similar to LACA, the $l$-th LASA adapter injects such spatial information explicitly through the self-attention mask $\mathbf{SM}^l$. Note that $\mathbf{SM}^l\in\mathbb{R}^{p_l^2\times p_l^2}$ is designed to specify whether the visual features should attend to one another or not. We omit the superscript $l$ in the following discussion.

\paragraph{Constructing self-attention mask from layout.} We propose to compose $\mathbf{SM}$ from the layout. First, we make an assumption that there are $K$ objects depicted by the layout. Note that objects that share the same noun entities are considered differently, for example, ``four apples and an orange'' leads to $K=5$. Second, we need to define the flattened visual index set $\mathcal{I}_k$ for each object $k$: 
\begin{align}
    \mathcal{I}_k=\{pi+j|\text{the}~(i,j)\text{-th visual feature belongs the object}~k\}\nonumber
\end{align} 
Then, we can compose $\mathbf{SM}$ using the following rules:

\begin{enumerate}
    \item if $i\in\cup_k\mathcal{I}_k$, then $\mathbf{SM}[i,j]=1$ if we have $i$ and $j$ assigned to the same object k, otherwise 0. Mathematically, the condition can expressed as $\sum_{k=1}^K\bone[i\in\mathcal{I}_k]\bone[j\in\mathcal{I}_k] > 0$
    \item if $i\notin\cup_k\mathcal{I}_k$, then we have $\mathbf{SM}[i,:]=1$.
\end{enumerate}

\begin{table}[t]
    \centering
    \begin{tabular}{lc}\hline
    Model &  \# Parameters (in billions)\\\hline
    Stable Diffuison & 1.06B\\\hline
    GLIGEN & 1.27B\\ 
    LACA & \textbf{1.12B}\\
    LACA+LASA &1.16B\\\hline
    \end{tabular}
    \caption{Model parameters.}
    \label{tab:params}
\end{table}
\begin{algorithm}[h]
  \caption{Construction of self-attention mask $\mathbf{SM}$.}
  \label{alg:self-att-mask}
  \small      
  \begin{algorithmic}
    \STATE \textbf{Input}: $K$ index sets $\mathcal{I}_1,\ldots,\mathcal{I}_K$, and an all-zeros tensor $\calM\in\mathbb{R}^{p^2\times p^2\times K}$.
    \STATE \textbf{[Composing self-attention mask for each object]}
    \FOR{$k=0, \ldots, K-1$}
    \STATE {Set $\calM[i,j,k]=1$, $\forall i,j\in\mathcal{I}_k$}
    \STATE {Set $\calM[i,:,k]=\bone$, $\forall i\notin\mathcal{I}_k$}
    \ENDFOR
    \STATE \textbf{[Reducing $\calM$ to $\mathbf{SM}$]}
    \STATE Intialize $\mathbf{SM} = \calM[:,:,0]$
    \FOR{$k=1, \ldots, K-1$}
    \STATE Obtain non-zero index set $\calI_{\emptyset,k}=\{i|\calM[i,:,k]\neq\bone\}$
    \STATE Obtain non-zero index set $\calI_{\emptyset,\mathbf{SM}}=\{i|\mathbf{SM}[i,:,k]\neq\bone\}$
    \STATE For $i\in\calI_{\emptyset,k}\cap \calI_{\emptyset,\mathbf{SM}}$, we set \\
    ~~~~$\mathbf{SM}[i,:] = \mathbf{SM}[i,:]~\text{or}~\calM[i,:,k]$
    \STATE For $i\in\calI_{\emptyset,k}\setminus(\calI_{\emptyset,k}\cap \calI_{\emptyset,\mathbf{SM}})$, we set \\
    ~~~~$\mathbf{SM}[i,:] = \calM[i,:,k]$
    \ENDFOR
    \STATE \textbf{Output:} Self-attention mask $\mathbf{SM}$
  \end{algorithmic}
\end{algorithm}

Here $\bone[\cdot]$ is an indicator function. The intuition of rule 1 is that a visual feature will aggregate information from all other visual features that share the same objects. Rule 2 allows non-object visual features to aggregate information from all the others. We highlight how to compose such an attention mask in Alg.\ref{alg:self-att-mask}.
\paragraph{Module designs.} Similar to LACA, we directly initialize LASA using the self-attention weight from the Stable Diffusion models. We only apply LASA to the low-resolution visual maps (the intermediate visual feature map with resolutions 16x16 and 8x8), which are more computationally affordable. We also add a zero convolution layer on top of LASA's output.

\paragraph{Composing LACA and LASA adapters.} For each transformer block, we add a LASA module in between the LACA module and the cross-attention module. We only add LASA to the last two downsampling blocks and the first two upsampling blocks.
\section{Additional Results}
\label{app:result}
\subsection{Ablation Study on CoT Variants}
The CoT strategy is directly linked to the performance of the generated layout. Following \S~\ref{sec:exp-cot}, we assess the effectiveness of different CoT variants by evaluating the performance of the LLM-generated layout. In particular, We report the object hit rate, mIoU, and the FID score of generated images in Table~\ref{tab:cot-variant}. The second variant, which we use in our major experiment, has shown superior performance over others. Surprisingly, variant 3, which does not separate each reasoning step explicitly, performs even worse than the one without CoT reasoning. 
\begin{table}[t]
    \centering
    \begin{tabular}{l|ccc}\hline
CoT strategy &  hit rate & mIoU &FID\\\hline
no CoT & 97.1\%&  19.84 & 31.95\\
variant 1 &\textbf{97.4}\%&  21.31 & 31.64\\
variant 2   &\textbf{97.4}\%&  \textbf{23.98} & \textbf{28.96}\\
variant 3   &96.9\%&  19.28 & 32.49\\
\hline
    \end{tabular}
    \caption{Layout performance of different CoT variants on Flickr30K.}
    \label{tab:cot-variant}
\end{table}
\subsection{Ablation Study on CFG Variants}
\label{sec:cfg-ablation}
We investigate the generative performance under different CFG guidances (Eq.~\ref{eq:cfg1}, Eq.~\ref{eq:cfg2} and Eq.~\ref{eq:cfg3}). We report the FID score on Flickr30K in Table~\ref{tab:cfg-guidance-fid}. Empirically, we observe that the employed CFG guidance works the best among others. We only incorporate layout modality into SD via the mentioned CFG guidances for the first 20\% denoising steps.

\begin{table}[t]
    \centering
    \begin{tabular}{l|c}\hline
CFG guidance &  FID\\\hline
Eq.~\ref{eq:cfg1} & \textbf{28.96}\\
Eq.~\ref{eq:cfg2} & 30.20\\
Eq.~\ref{eq:cfg3} & 29.19\\
\hline
    \end{tabular}
    \caption{Generative performance of different CFG guidance on Flickr30K}
    \label{tab:cfg-guidance-fid}
\end{table}
\begin{table}
    \centering
    \small
    \begin{tabular}{l|ccccccc}\hline
         \multirow{2}{*}{} & \multicolumn{5}{c}{Flick30K}\\
          &  \multirow{2}{*}{FID} & \multicolumn{3}{c}{GLIP score} & \multirow{2}{*}{GLIP rate}\\
         &&AP&$\text{AP}_{50}$&$\text{AP}_{75}$&\\\hline
        LACA & 28.96 & {58.5} &{68.1} &{59.4} & \textbf{83.3}\\
        LACA+LASA & \textbf{26.42} & \textbf{59.4} & \textbf{68.4} &\textbf{60.0} & \textbf{83.3}\\\hline
        \multirow{3}{*}{} & \multicolumn{5}{c}{COCO2017}\\
          &  \multirow{2}{*}{FID} & \multicolumn{3}{c}{GLIP score} & \multirow{2}{*}{GLIP rate}\\
         &&AP&$\text{AP}_{50}$&$\text{AP}_{75}$&\\\hline
        LACA & 23.28 & {58.8} & {67.3} & \textbf{{50.1}} & 78.1\\
        LACA+LASA& \textbf{22.13} & \textbf{58.4} & \textbf{68.9} & 49.7 & \textbf{78.2}\\\hline
    \end{tabular}
    \caption{Generative performance of LASA adapter}
    \label{tab:lasa-result}
\end{table}
\subsection{Generative Performance of LASA Adapter}
We validate the effectiveness of the LASA adapter on both the Flickr30K and COCO2017 datasets. This study involved a comparative analysis of the LACA adapter alone and its integration with the LASA adapter.  We use LLM-generated layouts to generate images and evaluate the FID score, GLIP score, and GLIP rate of the models. Table~\ref{tab:lasa-result} shows that the combination of LASA with LACA resulted in enhanced capabilities in producing more realistic images. Notably, there was an improvement of 2.54 and 1.15 in performance on the Flickr30K and COCO2017 datasets, respectively. The measures of layout accuracy and composition accuracy demonstrated that both adapters were comparably effective in integrating objects into images.  Although LASA contributes to higher-quality image generation, it also leads to increased sampling time. In speed tests conducted on an A100 GPU, the sampling time averaged 2.5 seconds for LACA but extended to 5.2 seconds when utilizing LACA combined with LASA.
\subsection{Visualizations}
\begin{figure*}
    \centering
    \begin{tabular}{c}
    \includegraphics[width=1\linewidth]{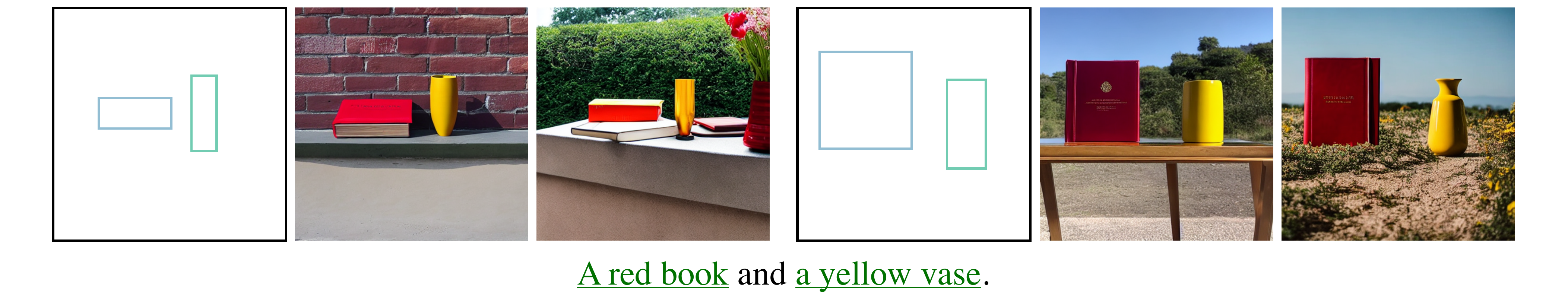}\\(a) \\
    \includegraphics[width=1\linewidth]{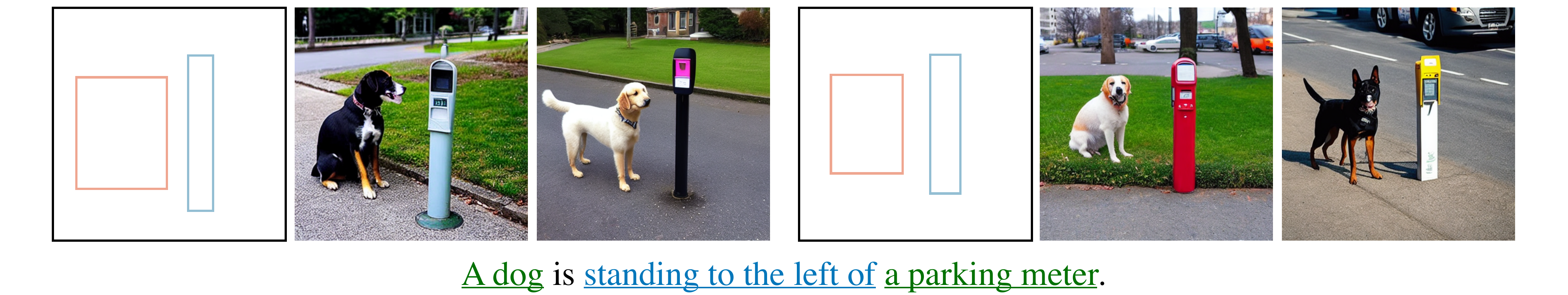}\\(b) \\
    \includegraphics[width=1\linewidth]{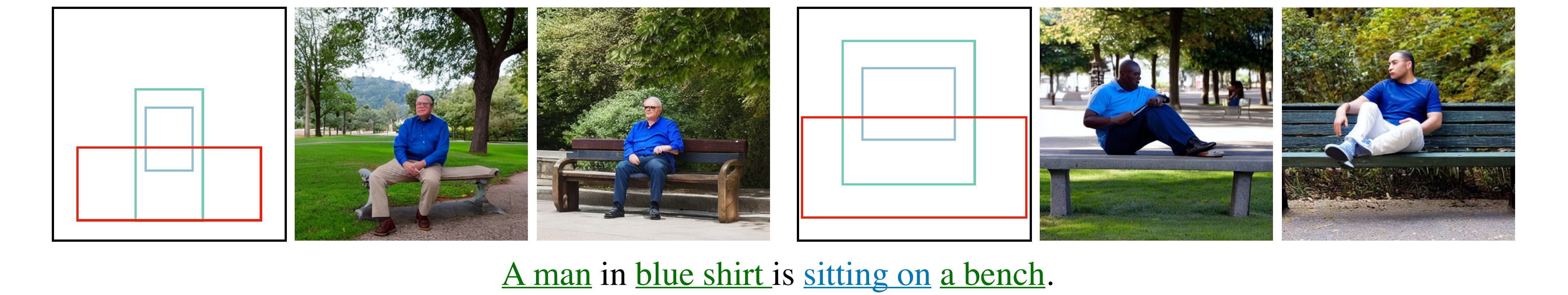}\\(c) \\
    \end{tabular}
    \caption{Additional visualization 1.}
    \label{fig:add-vis-1}
\end{figure*}

\begin{figure*}
    \centering
    \begin{tabular}{c}
    \includegraphics[width=1\linewidth]{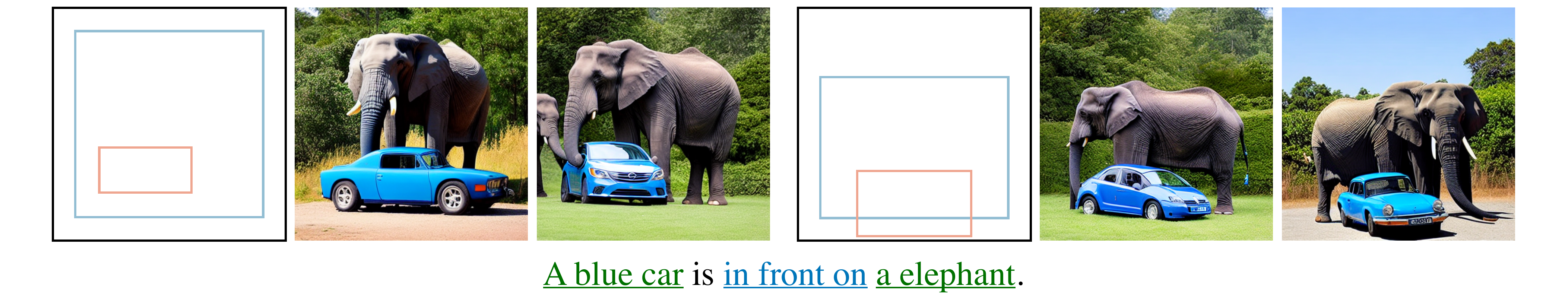}\\ (d) \\
    \includegraphics[width=1\linewidth]{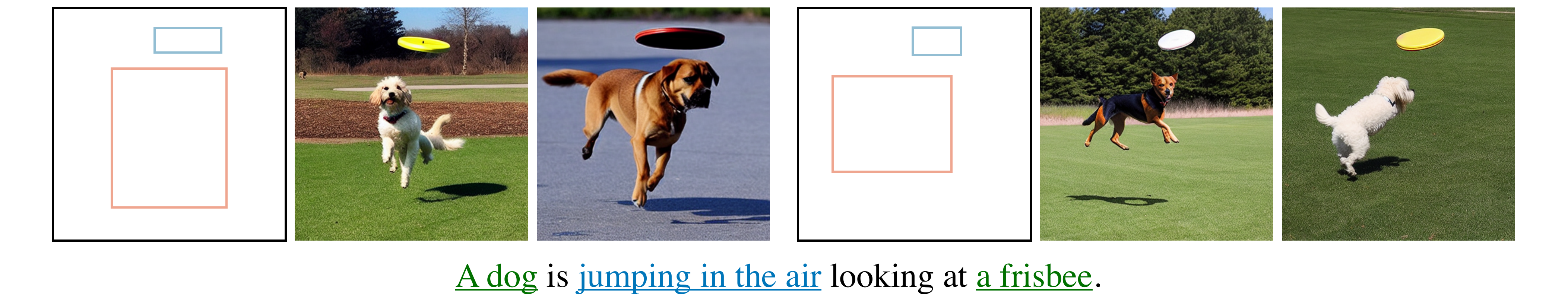}\\ (e) \\
    \includegraphics[width=1\linewidth]{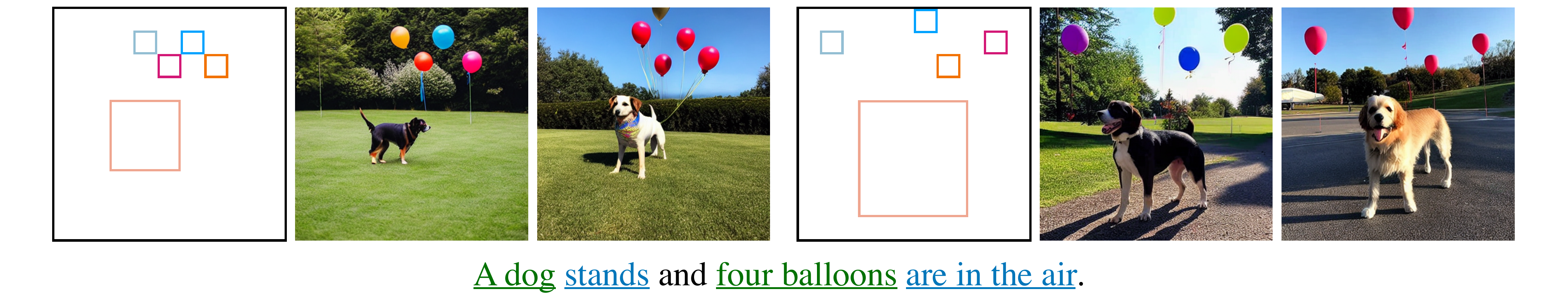} \\ (f)\\
    \end{tabular}
    \caption{Additional visualization 2.}
    \label{fig:add-vis-2}
\end{figure*}

We provide additional visualizations of the generated images in Fig.~\ref{fig:add-vis-1} and Fig.~\ref{fig:add-vis-2}. Specifically, we showcase 6 caption examples and their generated images. For each caption, we use an LLM to generate two layouts. Then we generate two images from the layout. As we can observe, our proposed method can generate reasonable layouts most of the time. Note that the generated objects do not necessarily lie within the given bounding boxes. We hypothesize this is because the layout information is only injected via LACA at the early stage of the denoising process. When LACA is no longer employed, the Stable Diffusion model takes its liberty to compose the objects. We believe this is beneficial since the generative error from the layout can be alleviated by the Stable Diffusion model, thereby achieving higher image quality. 


\end{document}